\documentclass{article}

\usepackage[final]{neurips_2021}

\usepackage[utf8]{inputenc} 
\usepackage[T1]{fontenc}    
\usepackage{hyperref}       
\usepackage{url}            
\usepackage{booktabs}       
\usepackage{amsfonts}       
\usepackage{nicefrac}       
\usepackage{microtype}      
\usepackage{xcolor}         

\usepackage{amsmath}

\newcommand{\indice}[1]{
\operatorname*{#1}
}

\usepackage{bbold}

\usepackage{colortbl}
\usepackage{subcaption,booktabs}

\usepackage{graphicx} 
\usepackage{wrapfig}

\usepackage[linesnumbered,ruled,vlined]{algorithm2e} 

\usepackage{pgffor}

\usepackage{pgf}

\usepackage{adjustbox}

\usepackage{comment}

\newcolumntype{P}[1]{>{\centering\arraybackslash}p{#1}}
\newcolumntype{M}[1]{>{\centering\arraybackslash}m{#1}}

\usepackage{tikz}
\usetikzlibrary{math}

\newcommand\styleAuthor{1}
\usepackage{ifthen}
\ifthenelse{\equal{\styleAuthor}{\string 1}}
{} 
{\usepackage{authblk}} 

\usepackage[nomessages]{fp}

\usepackage{amsthm}
\newtheorem{theorem}{Theorem}[section]

\theoremstyle{definition}

\usepackage{amsmath}
\DeclareMathOperator{\sigmoid}{sigmoid}
\DeclareMathOperator{\domain}{domain}

\title{
Stochastic Average Gradient : \\ 
A 
Simple
Empirical Investigation
}

\ifthenelse{\equal{\styleAuthor}{\string 1}}
{
    \author{%
      Pascal Junior Tikeng Notsawo \\ 
      \texttt{pascal.junior.tikeng.notsawo@umontreal.ca} \\
      \AND
      {\normalfont DIRO, Université de Montréal, Montréal, Quebec, Canada} \\
    }
} 
{
    \author[*]{Pascal Tikeng Notsawo}
    \author[*]{Guillaume Dumas}
    \author[*]{Author C}
    \author[**]{Author D}
    \author[**]{Author E}
    \affil[*]{Department of Computer Science, \LaTeX\ University}
    \affil[**]{Department of Mechanical Engineering, \LaTeX\ University}
    
} 

\begin{document}

\maketitle

\begin{abstract}

Despite the recent growth of theoretical studies and empirical successes of neural networks, gradient backpropagation is still the most widely used algorithm for training such networks. On the one hand, we have deterministic or full gradient (FG) approaches that have a cost proportional to the amount of training data used but have a linear convergence rate, and on the other hand, stochastic gradient (SG) methods that have a cost independent of the size of the dataset, but have a less optimal convergence rate than the determinist approaches. To combine the cost of the stochastic approach with the convergence rate of the deterministic approach, a stochastic average gradient (SAG) has been proposed. SAG is a method for optimizing the sum of a finite number of smooth convex functions. Like SG methods, the SAG method's iteration cost is independent of the number of terms in the sum. In this work, we propose to compare SAG to some standard optimizers used in machine learning. SAG converges faster than other optimizers on simple toy problems and performs better than many other optimizers on simple machine learning problems. We also propose a combination of SAG with the momentum algorithm and Adam. These combinations allow empirically higher speed and obtain better performance than the other methods, especially when the landscape of the function to optimize presents obstacles or is ill-conditioned 
\footnote{This work is reproducible at \url{https://github.com/Tikquuss/sag_torch}}.


\end{abstract}

\section{Introduction}
\label{introduction}

In many domains, several problems can be reduced to the minimization of the sum of a finite number 
of 
functions 
$$g = \frac{1}{n} \sum_{i=1}^n f_i$$
That is 
\begin{equation}
\label{eq:problem}
\indice{minimize}_{x \in \Omega \subset \mathbb{R}^p} \ \ g(x) = \frac{1}{n} \sum_{i=1}^n f_i (x)
\end{equation}

Gradient descent \citep{Cauchy2009ANALYSEM,bottou-98x, doi:10.1137/070704277,JMLR:v12:duchi11a,kingma2014adam} optimize such functions with a rule of the form :
$$x^{k+1} = x^k - \alpha_k D^k$$
where  $\alpha_k$ is the step size at iteration $k$; and $D^k$ a function of the past gradients $G_1, \dots, G_k$ of $g$ at $x^1, \dots, x^k$, respectively, or of the estimators of these gradients; such that $\mathbb{E}[D^k | x^{k-1}] = \nabla_{}g(x^k)$. More specifically, $G_k = \nabla_{}g(x^k)$ is the gradient of $g$ at $x^k$, the parameter update at time $k$ given the optimization algorithm of choice, and $\{ \alpha_k, k \ge 0 \}$ is a predefined deterministic sequence of positive real numbers such that $\sum_{k=1}^{\infty} \alpha_k = \infty$ and $\sum_{k=1}^{\infty} \alpha_k^2 < \infty$. The first of these two conditions is to make sure that the total displacement $\sum_{k=1}^{\infty} \alpha_k \nabla g(x^k)$ can be unbounded, so the optimal solution can be reached even if we start far away from it. The second condition (the finite sum of squares) is to decrease fast enough for the algorithm to converge.  For convex functions, gradient descent converges to a global minimum (if one exists).

Problem \ref{eq:problem} is very common in deep learning, where the goal is to minimize the regularized cost function 
$$\mathcal{J}(\theta) = \mathbb{E}_{s \sim F} [\ell(s, \theta)] + \lambda r(\theta) = \int \ell(s, \theta) dF(s) +  \lambda r(\theta) $$

where the function $\ell(s, \theta)$ measures how well the neural network with parameters $\theta$ predicts the label of a data sample $s$, $F$ is the cumulative distribution function of the data distribution, $r(\theta)$ is the regularizer (e.g. $\ell_2$-regularization $\frac{1}{2} \| \theta \|^2$), and $\lambda \in \mathbb{R}_{+}$ the regularization strength. In practice, $F$ is generally unknown, and the empirical distribution of a given dataset $\mathcal{D}$ is used. The regularized empirical risk obtained can be written as a sum of $|\mathcal{D}|$ functions 
$$\mathcal{J}(\theta) = \frac{1}{|\mathcal{D}|} \sum_{s \in \mathcal{D}} \big[ \ell(s, \theta) + \lambda r(\theta) \big]$$

This is the case, for example, of the least squares regression, with
$$\mathcal{D} = \{(x_i, y_i) \in \mathbb{R}^p \times  \mathbb{R} \}_{i=1}^n \text{ and } \ell((x, y), \theta) = \| x^T \theta - y \|^2_2$$
or the logistic regression where 
$\ell$ is the negative log-likelihoods \footnote{
The decision boundary is $x^T \theta = 0$, i.e. we want $x^T \theta > 0$ for $y=1$ and $x^T \theta < 0$ for $y=-1$, thas is $y x^T \theta > 0 \Longleftrightarrow \sigmoid (y x^T \theta) =  1/(1 + \exp(-y x^T \theta)) > 1/2$. To maximize $\sigmoid (y x^T \theta) \in [0, 1]$, we minimize $- \log \big( \sigmoid (y x^T \theta) \big) \in \mathbb{R}^+$, which gives our loss function.} :
$$\mathcal{D} = \{(x_i, y_i) \in \mathbb{R}^p \times \{-1, 1\} \}_{i=1}^n \text{ and } \ell((x, y), \theta) = \log( 1 + \exp(-y x^T \theta))$$ 



One of the challenges that gradient-based methods face in practice is the ill-conditioned surfaces, when the hessian of the function to optimize has some large positive eigenvalues (i.e. high-curvature directions) and some eigenvalues close to $0$ (i.e. low-curvature directions). In this case, vanilla gradient descent bounces back and forth in high curvature directions and slowly progresses in low curvature directions. In addition to these ill-conditioned surfaces, there are obstacles such as saddle points and critical surfaces (cliffs, valleys, plateaus, ravines, and other flat regions), extremely sharp or flat minima. 


The aim of this work is to empirically investigate the performance of stochastic average gradient (SAG) \citep{schmidt2013minimizing} on this type of problem. We limit ourselves for the first time on 
simple toys finite data
problems where each $f_i$ is smooth and convex, although in modern applications, $n$, the number of data points (or training examples) can be extremely large (e.g. datasets used to train large-scale deep learning models like GPT-3 \citep{brown2020language}), while there is often a large amount of redundancy between examples. In addition to this basic setting, we will also be interested in toys cases where the sum $g$ is strongly convex, with the use of a strongly-convex regularizer such as the squared $\ell_2$-norm, resulting in problems of the form :

\begin{equation}
\label{eq:problem_reg}
\indice{minimize}_{x \in \mathbb{R}^p} \ \ \frac{\lambda}{2} \| x \|^2 + \frac{1}{n} \sum_{i=1}^n f_i (x) = \frac{1}{n} \sum_{i=1}^n \Big[ \frac{\lambda}{2} \| x \|^2 + f_i (x)  \Big]
\end{equation}
    
The resulting function g will be strongly convex, provided that the individual  functions $f_i$ are convex.


We then extend our investigations to slightly more complex problems where we optimize deep neural networks on toys dataset. Many deep models are guaranteed to have an extremely large number of local minima. It has been proven that this is not necessarily a problem. Most local minima are of good quality (almost equivalent in cost to the global minimum) \citep{DBLP:conf/nips/DauphinPGCGB14}. The biggest obstacle to the optimization of $g$ in deep learning remains the presence of saddle points. In low dimensions (small $p$), local minima are more common, while in high dimensions, local minima are rare and saddle points more common.  Most of the training time is spent on traversing flat valleys of the Hessian matrix or circumnavigating tall mountains via an indirect arcing path, and the trajectory of traversing such flat valleys and circumventing such mountains may be long and result in excessive training time \citep{ChallengesSargur}.



The rest of the paper is organized as follows.
We define some terms used in our work in section \ref{sec:definitions}, then we present SAG in section \ref{sec:motivation}, the related works in section \ref{sec:related_works}, the convergence analysis and the implementation details in sections \ref{sec:convergence_analysis} and \ref{sec:implementation_details} respectively. We finally  present the experiments settings and the results in section ~\ref{sec:exp_results}, then summarise and conclude our work in section ~\ref{sec:conclusion}.


\section{Definitions}
\label{sec:definitions}

We assume $g : \mathbb{R}^p \to \mathbb{R}$ unless otherwise noted. The function $g$ is convex if for all $x, y \in \domain(g)$ and all $t \in [0, 1]$ 
$$g(tx + (1 - t) y) \le t g(x) + (1 - t) g(y)$$ 
or equivalently if for all $x, y \in \domain(g)$,
$$g(x) \ge g(y) + \nabla g(y)^T (x - y) $$
if $g$ is differentiable. If the inequality holds strictly (i.e. $<$ rather than $\le$) for all $t \in (0, 1)$ and $x \ne y$, then we say that $g$ is strictly convex, so strict convexity implies convexity.
Geometrically, convexity means that the line segment between two points on the graph of $g$ lies on or above the graph itself. If $g$ is convex, then any local minimum of $g$ in any convex set $X \subset \domain(g)$ is also a global minimum. 
Strict convexity means that the line segment lies strictly above the graph of $g$, except at the segment endpoints. If $g$ is strictly convex, then at most, one local minimum of $g$ in $X$ exists. Consequently, if it exists, it is the unique global minimum of $g$ in $X$ \footnote{\url{https://ai.stanford.edu/~gwthomas/notes/convexity.pdf}}. 

For $\mu > 0$, the function $g$ is $\mu$-strongly convex if the function $$x \mapsto g(x) - \frac{\mu}{2} \| x \|^2$$ is convex, or equivalently if for all $x, y \in \domain(g)$,
$$g(x) \ge g(y) + \nabla g(y)^T (x - y) + \frac{\mu}{2} \| x - y \|^2 $$
if $g$ is differentiable.
Strong convexity doesn’t necessarily require the function to be differentiable, and the gradient is replaced by the sub-gradient when the function is non-smooth. Intuitively speaking, strong convexity means a quadratic lower bound exists on the growth of the function. This directly implies that a strong convex function is strictly convex since the quadratic lower bound growth is, of course, strictly greater than the linear growth \footnote{\url{https://xingyuzhou.org/blog/notes/strong-convexity}}.

Let $G(x) = \nabla g(x) \in \mathbb{R}^{p}$ and $\mathcal{H}(x) = \nabla^2 g(x) \in \mathbb{R}^{p \times p}$ be respectively the gradient and the local hessian matrix of $g$ at $x$, assuming that $g$ is twice-differentiable. If $G(x) = 0$, then $x$ is a critical/stationary point of $g$. In this case, the determinant $d(x)$ of $\mathcal{H}(x)$ is equal to the Gaussian curvature of the surface of $g$ considered as a manifold. The eigenvalues of $\mathcal{H}(x)$ are the principal curvatures of the $g$ at $x$, and the eigenvectors are the principal directions of curvature.
If $d(x) > 0$, $x$ is a local  maximum of $g$ if $\mathcal{H}(x)$ is negative definite (all its eigenvalues are negative), and a local minimum of $g$ if $\mathcal{H}(x)$ is a positive definite (all its eigenvalues are positive).  Some local optimums can be very flat (i.e. there is a large enough neighbourhood of $x$ that contains only local optima) or sharp (the loss function near $x$ has a high condition number, i.e. very small perturbation of $x$ can cause large variation in $g$).
If $d(x) < 0$ (some eigenvalues are positive and others are negative), $x$ is a saddle point of $g$. If $d(x) = 0$ (there is at least one zero eigenvalue, i.e. $\mathcal{H}(x)$ is undefined), we can't conclude, and the point $x$ could be any of a minimum, maximum or saddle point.
If the hessian matrix of $g$ is positive semi-definite at any point of $\domain(g)$, then $g$ is convex and the point $x$ such that $G(x) = 0$ is its global minimum. If it is instead negative semi-definite at any point of $\domain(g)$, then $g$ is concave and the point $x$ such that $G(x) = 0$ is its global maximum.

\section{Motivation}
\label{sec:motivation}

Gradient descent \citep{bottou-98x} is one of the most popular algorithms to perform optimization and by far the most common way to optimize neural networks. FG method \citep{Cauchy2009ANALYSEM} uses iterations of the form
$$x^{k+1} = x^{k} - \alpha_k \nabla g(x^k) = x^{k} -  \frac{\alpha_k}{n} \sum_{i=1}^n \nabla f_i (x^k)$$

FG is generally called batch gradient descent in deep learning since it calculates the error for each example in the training dataset but only updates the model after all training examples have been evaluated. Therefore, its cost per iteration is $\mathcal{O}(n)$.

Assuming that a minimizer $x^*$ exists and $g$ is convex, then under standard assumptions, the sub-optimality achieved on iteration $k$ of the FG method with a constant step size is given by a sublinear convergence rate \citep{YuriiNesterovOptim2004,schmidt2013minimizing}
    
$$g(x^k) - g(x^*) = \mathcal{O} (1/k)$$

When $g$ is strongly convex, the error also satisfies  a linear convergence rate (also known as a geometric or exponential rate because a fixed fraction cuts the error on each iteration) \citep{YuriiNesterovOptim2004,schmidt2013minimizing}
$$g(x^k) - g(x^*) =  \mathcal{O}(\rho^k) \text{ for some } \rho < 1$$
    
This $\rho$ depends on the condition number of $g$, i.e. on how sensitive the output of $g$ is on its input \footnote{$L/\mu$  (change in output = condition number $\times$ change in input)}. One drawback of the FG approach is that it requires computing all the gradients at each iteration, which can be tedious when $n$ is very large.

The basic SG method for optimizing \ref{eq:problem} uses iterations of the form
$$x^{k+1}  = x^{k} - \alpha_k \nabla f_{i_k} (x^k)$$

where at each iteration an index $i_k$ is sampled uniformly from the set $\{1, \dots, n\}$. The randomly chosen gradient $\nabla f_{i_k} (x^k)$ yields an unbiased estimate of the true gradient $\nabla g(x^k)$ :
$$\mathbb{E}_{i_k \sim \mathcal{U}(\{1, \dots, n\})}[\nabla f_{i_k} (x^k)] = \frac{1}{n} \sum_{i=1}^n \nabla f_i (x^k) = \nabla g(x^k)$$

Under standard assumptions and for a suitably chosen decreasing step-size sequence $\{ \alpha_k, k \ge 0 \}$  \citep{doi:10.1137/070704277,schmidt2013minimizing}, the SG iterations have an expected sub-optimality for convex objectives of
$$\mathbb{E}[g(x^k)] - g(x^*) = \mathcal{O}(1/\sqrt{k})$$
and an expected sub-optimality for strongly-convex objectives of
$$\mathbb{E}[g(x^k)] - g(x^*) = \mathcal{O}(1/k)$$

These sublinear rates are slower than the corresponding rates for FG. Under certain assumptions, these convergence rates are optimal in a model of computation where the algorithm only accesses the function through unbiased measurements of its objective and gradient. Thus, we should not expect to be able to obtain the convergence rates of the FG method if the algorithm only relies on unbiased gradient measurements. Can we have one gradient per iteration and achieve the same rate as FG?

Mini-batch gradient descent is a variation of the SG algorithm that splits the training dataset into small batches used to calculate model error and update model coefficients. In other words, we select a batch $\mathcal{B} \subset \{1, \dots, n\}$ randomly at each iteration and do the update as follows:

$$x^{k+1} = x^{k} - \frac{\alpha_k}{|\mathcal{B}|} \sum_{i \in \mathcal{B}} \nabla f_i (x^k) $$

But this allows to make a trade-off between the cost per iteration and the convergence rate: either we choose $\mathcal{B}$ is too big, and we get a better rate and a big cost of $\mathcal{O}(|\mathcal{B}|)$ per iteration, or we choose $\mathcal{B}$ so that $|\mathcal{B}|$ is too small, and we get a lower rate and a cost in $\mathcal{O}(1)$ per iteration.

The SAG iterations take the form
$$x^{k+1} = x^{k} -  \frac{\alpha_k}{n} \sum_{i=1}^n y_i^k$$
where at each iteration a random index $i_k$ is selected (not necessarily uniformly from $\{1, \dots, n\}$ as we will see below) and we set
$$
y_i^k = \left\{
    \begin{array}{ll}
        \nabla f_i (x^k) & \mbox{if } i = i_k \\
        y_i^{k-1} & \mbox{otherwise.}
    \end{array}
\right.
$$

Like the FG method, the step incorporates a gradient with respect to each function. But, like the SG method, each iteration only computes the gradient with respect to a single example and the cost of the iterations is independent of $n$ : we take a step in the direction of the average of $y_i^k$.

With the mini-batch version of SAG, the update becomes 
$$
y_i^k = \left\{
    \begin{array}{ll}
        \nabla f_i (x^k) & \mbox{if } i \in \mathcal{B} \\
        y_i^{k-1} & \mbox{otherwise.}
    \end{array}
\right.
$$

\section{Related works}
\label{sec:related_works}

In the following $D^k$ is a function of the past gradients $G_1, \dots, G_k$ of $g$ at $x^1, \dots, x^k$, respectively, or of the estimators of these gradients. In the papers introducing these algorithms, $D^k = G_k$ in general, i.e. $D^k$ is deterministic. But their SG version can be developed with $D^k = \nabla f_{i_k} (x^k)$ for a randomly sampled $i_k \in \{1, \dots, n\}$, or their mini-batch version with $D^k = \frac{1}{|\mathcal{B}|} \sum_{i \in \mathcal{B}} \nabla f_i (x^k)$ for a random sample $\mathcal{B} \subset \{1, \dots, n\}$, or their SAG version with an appropriate choice of past gradients to use and how to use them.


SG methods that incorporate a each iteration $k$ a momentum term $m^k = x^k - x^{k-1} = - \alpha_{k-1} D^{k-1}$ use iterations of the form \citep{POLYAK19641, sutton:problems}
$$x^{k+1} = x^k - \alpha_k D^k + \beta_k m^k$$
It is common to set all $\beta_k = \beta_1$ for some constant $\beta_1 \in [0, 1)$, and in this case, we can rewrite the SG with momentum \citep{doi:10.1137/S1052623495294797}  method as 
$$x^{k+1} = x^k - \sum_{j=0}^k \alpha_j \beta_1^{k-j} D^j$$

The momentum algorithm accumulates an exponentially decaying moving average of past gradients and continues to move in their direction. Formally, the momentum algorithm introduces a variable $v$ that plays the role of velocity: the direction and speed at which the parameters move through parameter space. The hyperparameter $\beta_1$ determines how quickly the contributions of previous gradients exponentially decay. The above update rule can be rewritten in terms of the velocity as  ($v^0 = 0$):
$$v^{k+1} = \beta_1  v^{k} - \alpha_k D^k$$
$$x^{k+1} = x^{k} + v^{k+1}$$
Since we have with this
$$
v^{k+1} = - \sum_{j=0}^k \alpha_j \beta_1^{k-j} D^j
$$
The SAG version of momentum becomes 
$$x^{k+1} = x^{k} + \frac{\alpha_k}{n} \sum_{i=1}^n y_i^k $$
where at each iteration, a random index $i_k$ is selected,  and we set
$$
y_i^k = \left\{
    \begin{array}{ll}
        v_i^{k+1} = \beta_1  v_i^{k} - \alpha_k D_i^k & \mbox{if } i = i_k, \text{ with } D^k = \nabla f_{i_k} (x^k) \\
        y_i^{k-1} & \mbox{otherwise.}
    \end{array}
\right.
$$

Nesterov accelerated gradient or Nesterov momentum \citep{Nesterov1983AMF, pmlr-v28-sutskever13} is a variant of the momentum algorithm that use an interim update $\tilde{x}^k =  x^{k} + \beta_1  v^{k}$ to compute de gradient $D^k$ at each iteration. That is : 

$$\tilde{x}^k =  x^{k} + \beta_1  v^{k} $$
$$\tilde{D}^k = \nabla g (\tilde{x}^k)$$
$$v^{k+1} = \beta_1  v^{k} - \alpha_k \tilde{D}_k$$
$$x^{k+1} = x^{k} + v^{k+1}$$



The AdaGrad algorithm \citep{10.5555/1953048.2021068}, individually adapts the learning rates of all model parameters by scaling them inversely proportional to the square root of the sum of all of their historical squared values. The update rule of AdaGrad is given by ($r^0 = 0$, $r^k$ accumulates squared gradient, division and square root are applied element-wise, $\epsilon$ is a very small number used to avoid divisions by $0$) :
$$r^{k+1} = r^{k} + D^k \odot D^k$$ 
$$x^{k+1} = x^{k} - \frac{\alpha_k}{\sqrt{r^{k+1}} + \epsilon} \odot D^k $$


The RMSProp algorithm \citep{HintonG2012} modifies AdaGrad to perform better in the non-convex setting by changing the gradient accumulation into an exponentially weighted moving average. RMSProp uses an exponentially decaying average to discard history from the extreme past to converge rapidly after finding a convex bowl as if it were an instance of the AdaGrad algorithm initialized within that bowl. Compared to AdaGrad, using the moving average introduces a new hyperparameter, $\beta_2 \in (0, 1]$, that controls the length scale of the moving average. The step of squared gradient accumulation is modified as follows: 
$$r^{k+1} = \beta_2 r^{k} + (1 - \beta_2) D^k \odot D^k$$ 


Adadelta \citep{zeiler2012adadelta} is an extension of Adagrad and RMSProp that seeks to reduce its aggressive, monotonically decreasing learning rate. Instead of accumulating all past squared gradients, Adadelta restricts the window of accumulated past gradients to some fixed size ($u^0 = 0$).

$$r^{k+1} = \beta_2 r^{k} + (1 - \beta_2) D_k \odot D_k$$ 
$$\Delta^{k+1}  = \frac{\sqrt{u^{k} + \epsilon}}{\sqrt{r^{k+1} + \epsilon}}$$
$$u^{k+1} = \beta_2 u^{k} + (1 - \beta_2) \Delta^{k+1}$$
$$x^{k+1} = x^{k} - \alpha_k \Delta^{k+1}$$


Adam \citep{kingma2017adam} is a combination of RMSProp and momentum. First, in Adam, momentum is incorporated directly as an estimate of the gradient's first-order moment (with exponential weighting). Second, Adam includes bias corrections to the estimates of both the first-order moments (the momentum term) and the (uncentered) second-order moments to account for their initialization at the origin.

$$ \tilde{x}^k =  x^{k} + \beta_1  v^{k} $$
$$\tilde{D}^k = \nabla g (\tilde{x}^k)$$
$$r^{k+1} = \beta_2 r^{k} + (1 - \beta_2) \tilde{D}^k \odot \tilde{D}^k$$ 
$$v^{k+1} = \beta_1  v^{k} - \frac{\alpha_k}{\sqrt{r^{k+1}} + \epsilon} \odot \tilde{D}^k$$
$$x^{k+1} = x^{k} + v^{k+1}$$

The most common Adam iteration update is written in term of momentum as 

\begin{equation*}
    \begin{split}
    & m^k = \beta_1 m^{k-1} + (1 - \beta_1) D^{k} \\
    & r^k = \beta_2 r^{k-1} + (1 - \beta_2)  D^{k} \odot D^{k}  \\
    & x^k =  x^{k-1} - \frac{\alpha_k}{\sqrt{r^k} + \epsilon} \odot m^k
    \end{split}
\end{equation*}


Adamax \citep{kingma2014adam,kingma2017adam} is a variant of Adam based on infinity norm. 

\begin{equation*}
    \begin{split}
    & m^k = \beta_1 m^{k-1} + (1 - \beta_1)  D^{k} \\
    & u^k = \max(\beta_2 u^{k-1}, |D^{k}| + \epsilon)  \\
    & x^k =  x^{k-1} - \frac{\alpha_k}{ (1 - \beta_1^k) u^k } \odot m^k
    \end{split}
\end{equation*}

AMSGrad \citep{reddi2018convergence} is a version of Adam that keeps a running maximum of the squared gradients instead of an exponential moving average.

\begin{equation*}
    \begin{split}
    & m^k = \beta_1 m^{k-1} + (1 - \beta_1)  D^{k} \\
    & \tilde{m}^{k} = \max(\tilde{m}^{k-1}, m^{k}) \\
    & r^k = \beta_2 r^{k-1} + (1 - \beta_2)  D^{k} \odot D^{k}  \\
    & x^k =  x^{k-1} - \frac{\alpha_k}{\sqrt{r^k} + \epsilon} \odot \tilde{m}^k
    \end{split}
\end{equation*}


All Adaptive methods can be summarized as follows \citep{défossez2020simple}. As hyper-parameters, we have $0 \le \beta_1 < \beta_2 \le 1$, and a non negative sequence $(\alpha_k)_{k \in \mathbb{N}^*}$. We define three vectors $m_k, r_k, x_k \in \mathbb{R}^p$ iteratively. Given $x^0 \in \mathbb{R}^p$ as our starting point, $m^0 = 0$, and $r^0 = 0$, we define for all iterations $k \in \mathbb{N}^*$
\begin{equation*}
    \begin{split}
    & m_i^k = \beta_1 m_i^{k-1} + D^{k}_i \\
    & r_i^k = \beta_2 r_i^{k-1} + \big( D^{k}_i \big)^2 \\
    & x_i^k =  x_i^{k-1} - \alpha_k \frac{m_i^k}{\sqrt{r_i^k} + \epsilon}
    \end{split}
\end{equation*}
    
.

The parameter $\beta_1$ is a heavy-ball style momentum parameter. The parameter $\beta_2$ controls the decay rate of the per-coordinate exponential moving average of the squared gradients. Taking $\beta_1 = 0$, $\beta_2 = 1$ and $\alpha_k = \alpha$ gives Adagrad \citep{JMLR:v12:duchi11a}. The original Adam algorithm \citep{kingma2014adam} uses a weighed average, rather than a weighted sum :
$$
\tilde{m}_i^k = (1 - \beta_1) \sum_{j=1}^k \beta_1^{k-j} D_i^{j-1} = (1 - \beta_1) m_i^k
$$
We can achieve the same definition by taking $\alpha_{adam} = \alpha \cdot \frac{1-\beta_1}{\sqrt{1-\beta_2}}$, since 
$$
\frac{\tilde{m}_i^k}{\sqrt{\tilde{r}_i^k}} =
\frac{1-\beta_1}{\sqrt{1-\beta_2}} 
\frac{m_i^k}{\sqrt{r_i^k}}
$$
with
$$\tilde{r}_i^k = (1 - \beta_2) r_i^k \text{ and } \tilde{m}_i^k = (1 - \beta_1) m_i^k$$

The original Adam algorithm further includes two corrective terms to account for the fact that $m^k$ and $r^k$ are biased towards $0$ for the first few iterations. Those corrective terms are equivalent to taking a step-size $\alpha_k$ of the form

$$ \alpha_{k, adam} = \alpha \cdot \frac{1-\beta_1}{\sqrt{1-\beta_2}} 
\cdot 
\overbrace{\frac{1}{\sqrt{1-\beta_1^k}}}^{ \text{ corrective term for } m^k }
\cdot 
\underbrace{\sqrt{1-\beta_2^k}}_{ \text{ corrective term for } r^k }$$

Early work on adaptive methods (e.g. \citep{mcmahan2010adaptive}) showed that Adagrad achieves an optimal rate of convergence of $\mathcal{O}(1/\sqrt{k})$ for convex optimization. \citet{ward2020adagrad} proved that Adagrad converges to a critical point for non convex objectives with a rate $\mathcal{O}(ln(k)/\sqrt{k})$ when using a scalar adaptive step-size. \citet{défossez2020simple} show a rate of $\mathcal{O}(p \ ln(k)/\sqrt{k})$ for Adam, and show that in expectation, the squared norm of the objective gradient averaged over the trajectory has an upper-bound which is explicit in the constants of the problem, parameters of the optimizer, the dimension $p$, and the total number of iterations $k$.

\section{SAG convergence rate}
\label{sec:convergence_analysis}

We assume that each function $f_i$ in (\ref{eq:problem}) is convex and differentiable (this makes $g$ also convex and differentiable), and that each gradient $\nabla f_i$ is Lipschitz-continuous with constant $L_i$, meaning that for all $x$ and $y$ in $\mathbb{R}^p$ and each $i$ we have
\begin{equation}
\label{eq:Lipschitz-continuous}
\| \nabla f_i(x) - \nabla f_i (y) \| \le L_i \| x - y \|
\end{equation}

This makes $\nabla g$ also Lipschitz-continuous with any constant $L \ge \frac{1}{n} \sum_{i=1}^n L_i$, like $ \indice{max}_{i}L_i$. Also, each gradient $\nabla f_i$ is Lipschitz-continuous with constant $L \ge \indice{max}_{i} L_i$. This is a fairly weak assumption on the $f_i$ functions, and in cases where the $f_i$ are twice-differentiable it is equivalent to saying that the eigenvalues of the hessians of each $f_i$ are bounded above by $L$. We will also assume the existence of at least one minimizer $x^*$ that achieves the optimal function value.

In addition to the above basic convex case, we will also consider the case where the average function $g = \frac{1}{n} \sum_{i=1}^n f_i$ is strongly-convex with constant $\mu > 0$, meaning that the function $x \mapsto g(x) - \frac{\mu}{2} \| x \|^2$ is convex. For twice-differentiable $g$, this is equivalent to requiring that the eigenvalues of the hessian of $g$ are bounded below by $\mu$. This is a stronger assumption that is often not satisfied in practical applications. Nevertheless, in many applications we are free to choose a regularizer of the parameters, and thus we can add an $\ell_2$-regularization term as in (\ref{eq:problem_reg}) to transform any convex problem into a strongly-convex problem (in this case we have $\mu \ge \lambda$). Note that strong-convexity implies the existence of a unique $x^*$ that achieves the optimal function value.

Let $\bar{x}_k = \frac{1}{k} \sum_{i=0}^{k-1} x^i$ be the average iterate and $\sigma^2 = \frac{1}{n} \sum_{i=1}^{n} \| \nabla f_i (x^*) \|$ the variance of the gradient norms at the optimum $x^*$.The convergence results consider two different initializations for the $y_i^0$ variables: 
\begin{itemize}
\item setting $y_i^0 = 0$  for all $i$
\item or setting them to the centered gradient at the initial point $x^0$ : $y_i^0 = \nabla f_i (x^0) - \nabla g (x^0)$
\end{itemize}
The convergence results are expressed in terms of expectations $\mathbb{E}$ with respect to the internal randomization of the algorithm (the selection of the random variables $i_k$), and not with respect to the data which is assumed to be deterministic and fixed.  The $L$ we use in the following is a Lipschitz-continuous constant common to all $\nabla f_i$, as $\indice{max}_{i} L_i$.

\begin{theorem}
With a constant step size of $\alpha = \frac{1}{16L}$, the SAG iterations satisfy for $k \ge 1$ :
\begin{equation}
\label{eq:part1theorem1}
\mathbb{E}[g(\bar{x}^k)] - g(x^*) \le \frac{32n}{k} C_0 \in \mathcal{O}\bigg(\frac{1}{k}\bigg)
\end{equation}
where if we initialize with $y_i^0 = 0$ for all $i$ we have
$$C_0 = g(x^0) - g(x^*) +  \frac{4L}{n} \| x^0 - x^* \|^2 + \frac{\sigma^2}{16L}$$
and if we initialize with $y_i^0 = \nabla f_i (x^0) - \nabla g (x^0)$ for all $i$ we have
$$C_0 = \frac{3}{2} \big[ g(x^0) - g(x^*) \big] +  \frac{4L}{n} \| x^0 - x^* \|^2$$

Further, if g is $\mu$-strongly convex we have
$$\mathbb{E}[g(x^k)] - g(x^*) \le \bigg( 1 - min \Big\{ \frac{\mu}{16L}, \frac{1}{8n} \Big\} \bigg)^k C_0 \in \mathcal{O}\Bigg( \bigg( 1 - min \Big\{ \frac{\mu}{16L}, \frac{1}{8n} \Big\} \bigg)^k \Bigg)$$
\end{theorem}

The proof of this theorem is given in \citet{schmidt2013minimizing} [Appendix B] and involves finding a Lyapunov function for a non-linear stochastic dynamical system defined on the $y_i^k$ and $x_k$ variables that converges to zero at the above rates, and showing that this function dominates the expected sub-optimality $\mathbb{E}[g(x^k)] - g(x^*)$. The equation (\ref{eq:part1theorem1}) is stated for the average $\bar{x}^k$, with a trivial change to the proof technique, but it can be shown to also hold for any iterate $x^k$ where $g(x^k)$ is lower than the average function value up to iteration $k$, $\frac{1}{k} \sum_{i=0}^{k-1} g(x^i)$. Thus, in addition to $\bar{x}^k$ the result also holds for the best iterate.

The bounds are valid for any $L$ greater than or equal to the minimum $L$ satisfying (\ref{eq:Lipschitz-continuous}) for each $i$, implying an $\mathcal{O}(1/k)$ and linear convergence rate for any $\alpha \le 1/16L$, but the bound becomes worse as $L$ grows. Although initializing each $y_i^0$ with the centered gradient may have an additional cost and slightly worsens the dependency on the initial sub-optimality $(g(x^0) - g(x^*))$, it removes the dependency on the variance $\sigma^2$ of the gradients at the optimum.

While the theorem is stated in terms of the function values, in the $\mu$-strongly-convex case we also obtain a convergence rate on the iterates because we have
$$
\frac{\mu}{2} \| x^k - x^* \|^2 \le g(x^k) - g(x^*)
$$

The SAG iterations have a worse constant factor because of the dependence on $n$. An appropriate choice of $x^0$ can improve the dependence on $n$ : we can set $x^0$ to the result of $n$ iterations of an appropriate SG method. In this setting, the expectation of $g(x^0) - g(x^*)$ is $\mathcal{O}(1/\sqrt{n})$ in the convex setting, while both $g(x^0) - g(x^*)$ and $\| x^0 - x^* \|^2$ would be in $\mathcal{O}(1/n)$  in the strongly-convex setting. 

If we use this initialization of $x^0$ and set $y_i^0 = \nabla f_i(x^0) - \nabla g(x^0)$, then in terms of $n$ and $k$ the SAG convergence rates take the form $\mathcal{O}(\sqrt{n}/k)$ and $\mathcal{O}(\rho^k/n)$ in the convex and strongly-convex settings, instead of the $\mathcal{O}(n/k)$ and $\mathcal{O}(\rho^k)$ rates implied by the theorem.

An interesting consequence of using a step-size of $\alpha = 1/16L$ is that it makes the method adaptive to the strong-convexity constant $\mu$. For problems with a higher degree of local strong-convexity around the solution $x^*$, the algorithm will automatically take advantage of this and yield a faster local rate. This can even lead to a local linear convergence rate if the problem is strongly-convex near the optimum but not globally strongly-convex. This adaptivity to the problem difficulty is in contrast to SG methods whose sequence of step sizes typically depend on global constants and thus do not adapt to local strong-convexity. \textit{We will test this on the Rosenbrock function in log scale, for which the SG method turns indefinitely around the global minimum and never reaches it}.

\section{SAG implementation Details}
\label{sec:implementation_details}

\citet{schmidt2013minimizing} discuss modifications that lead to better practical performance than this basic algorithm, including ways to reduce the storage cost, how to handle regularization, how to set the step size, using mini-batches, and using non-uniform sampling. 

\begin{algorithm}[H]
\caption{Basic SAG method for minimizing $\frac{1}{n} \sum_{i=1}^n f_i (x)$ with step size $\alpha$}
\label{sedat}
\SetAlgoLined
\DontPrintSemicolon
\Begin{
    $d = 0$ \CommentSty{$\ /*$ $d$ is use to track the quantity $\sum_{i=1}^n y_i$ $ */$}\;
    $y_i = 0$ for $i = 1, 2, \dots, n$\;
    \For{$k = 0, 1, \dots$}{
        Sample $i$ from $\{1, 2, \dots, n\}$\; 
        $d = d - y_i + \nabla f_i(x)$\;  
        $y_i = \nabla f_i(x)$\;
        $x = x - \frac{\alpha}{n} d$\;
    }
}
\end{algorithm}

\paragraph{Re-weighting on early iterations}
The more logical normalization is to divide $d$ by $m$, the number of data points that we have seen at least once (which converges to $n$ once we have seen the entire data set), when $y_i^0 = 0$
$$x = x - \frac{\alpha}{m} d$$

\paragraph{Exact and efficient regularization}
$$x = x - \alpha \bigg( \frac{d}{m} + \lambda x \bigg) = (1-\alpha \lambda) x - \frac{\alpha}{m} d$$

\paragraph{Mini-batches for vectorized computation and reduced storage}

$$x^{k+1} = x^{k} -  \frac{\alpha_k}{n} \sum_{i=1}^n y_i^k
\text{ with }
y_i^k = \left\{
    \begin{array}{ll}
        \nabla f_i (x^k) & \mbox{if } i \in \mathcal{B} \\
        y_i^{k-1} & \mbox{otherwise.}
    \end{array}
\right.
$$

\paragraph{ Structured gradients and just-in-time parameter updates}

For many problems the storage cost of $\mathcal{O}(np)$ for the $y_i^k$ vectors is prohibitiven but we can often use the structure of the gradients $\nabla f_i$ to reduce this cost. For example, let consider a linearly-parameterized model of the form 
\begin{equation}
\label{eq:problem1}
\indice{minimize}_{x \in \Omega \subset \mathbb{R}^p} \ \ g(x) = \frac{1}{n} \sum_{i=1}^n f_i (a_i^T x)
\end{equation}
    
Since each $a_i$ is constant, for these problems we only need to store the scalar $\nabla f_{i_k} (u_i^k)$ for $u_i^k = a_{i_k}^T x$ rather than the full gradient $a_i \nabla f_{i} (u_i^k)$. This reduces the storage cost from $\mathcal{O}(np)$ down to $\mathcal{O}(n)$. Examples of linearly-parameterized models include the least-squares regression 
\footnote{
$\ell(s = (x, y), \theta) 
= h(x^T \theta) \textit{ with } h(z) = (z - y)^2$
}, the logistic regression 
\footnote{
$\ell(s = (x, y), \theta)  = h(x^T \theta)  \textit{ with } h(z) = log( 1 + exp(-y z))$
}, feed forward neural networks, etc.

\section{Experiments settings Results }
\label{sec:exp_results}

We will use the following acronyms to designate our algorithms :

\begin{itemize}
    \item sgd : vanilla SGD 
    \item momentum : SGD with momentum  \citep{POLYAK19641, sutton:problems,doi:10.1137/S1052623495294797}
    \item nesterov : Nesterov Accelerated SGD \citep{Nesterov1983AMF, pmlr-v28-sutskever13}
    \item asgd : Averaged SGD proposed by \citet{10.1137/0330046} 
    \item rmsprop : RMSProp  \citep{HintonG2012} 
    \item rmsprop\_mom : RMSProp with momentum 
    \item rprop : resilient backpropagation algorithm \citep{298623}
    \item adadelta : Adadelta \citep{zeiler2012adadelta}
    \item adagrad : Adagrad \citep{JMLR:v12:duchi11a}
    \item adam : Adam \citep{kingma2014adam,kingma2017adam}
    \item amsgrad : AMSGrad  \citep{reddi2018convergence} 
    \item adamax  : Adamax \citep{kingma2014adam}
    \item custom\_adam : custom adam algorithm without amsgrad and that include the two corrective terms for $m^k$ and $r^k$
    \item adam\_inverse\_sqrt : Adam that decays the learning rate based on the inverse square root of the update number. It also supports a warmup phase where the learning rate is linearly increase  from some initial learning rate ($warmup\_init\_lr$) until the configured learning rate ($lr$). Thereafter, the learning rate is decay proportional to the number of updates, with a decay factor set to align with the configured learning rate.
    \begin{itemize} 
        \item  During warmup:
        $$lrs = linspace(start = warmup\_init\_lr, end = lr, steps = warmup\_updates)$$
        $$lr = lrs[step]$$
        \item After warmup:
        $$lr = \frac{decay\_factor}{\sqrt{update\_num}} 
        \text{ where }
        decay\_factor = lr * sqrt(warmup\_updates)$$
    \end{itemize}
    \item adam\_cosine : Adam that assign learning rate based on a cyclical schedule that follows the cosine function \citep{loshchilov2016sgdr}. It also supports a warmup phase where the learning rate is linearly increase  from some initial learning rate ($warmup\_init\_lr$) until the configured learning rate ($lr$). Thereafter, the learning rate is decay proportional to the number of updates, with a decay factor set to align with the configured learning rate.
    \begin{itemize} 
        \item  During warmup:
        $$lrs = linspace(start = warmup\_init\_lr, end = lr, steps = warmup\_updates)$$
        $$lr = lrs[step]$$
        \item After warmup:
        $$lr = lr\_min + 0.5*(lr\_max - lr\_min)*(1 + cos(t\_curr / t\_i))$$
        where $t\_curr$ is current percentage of updates within the current period range and $t\_i$ is the current period range, which is scaled by $t\_mul$ after every iteration.
    \end{itemize}
    \item sag : SAG \citep{schmidt2013minimizing}
    \item sag\_sgd : combinaition of SAG and momentum SGD with 
$$
y_i^k = \left\{
    \begin{array}{ll}
        v^{k+1} = \beta_1  v_i^{k} - \alpha_k D_i^k & \mbox{if } i = i_k, \text{ with } D^k = \nabla f_{i_k} (x^k) \\
        y_i^{k-1} & \mbox{otherwise.}
    \end{array}
\right.
$$ 
    \item sag\_adam : combinaition of SAG and Adam with
$$
y_i^k = \left\{
    \begin{array}{ll}
         \frac{m_i^k}{\sqrt{r_i^k} + \epsilon} & \mbox{if } i = i_k\\
        y_i^{k-1} & \mbox{otherwise.}
    \end{array}
\right.
$$ 
\end{itemize}

\subsection{Test functions for optimization}

\subsubsection{Rosenbrock function}


The vanilla rosenbrok function is given by $g_n(x) = \sum_{i=1}^{n/2} \big[ 100 (x_{2i} - x_{2i-1}^2)^2 + (x_{2i-1} - 1)^2 \big]$, with the gradient $\nabla_i g_n(x) =200 (x_i - x_{i-1}^2) \cdot \mathbb{1}_{i \in 2 \mathbb{N}} - \big[ 400 x_i (x_{i+1} - x_i^2) - 2(x_i - 1) \big] \cdot \mathbb{1}_{i \in  2 \mathbb{N} - 1}$, and $x^* \in \{ (1, \dots, 1), (-1, 1, \dots, 1) \} \subset \{ x, \nabla g_n(x) = 0 \}$
\footnote{
When the coordinates range from $0$ to $n-1$, $g_n(x) = \sum_{i=0}^{n/2-1} \big[ 100 (x_{2i+1} - x_{2i}^2)^2 + (x_{2i} - 1)^2 \big]$ and $\nabla_i g_n(x) = 200(x_{i} - x_{i-1}^2) \cdot \mathbb{1}_{i \in 2 \mathbb{N} + 1} - \big[ 400 x_i (x_{i+1} - x_i^2) - 2(x_i - 1) \big] \cdot \mathbb{1}_{i \in 2 \mathbb{N}}$.}. 
A more involved variant is given by $g_n(x) = \sum_{i=1}^{n-1} \big[ 100 (x_{i+1} - x_i^2)^2 + (x_i - 1)^2 \big]$, with the gradient $\nabla_i g_n(x) = 200 (x_i - x_{i-1}^2) \cdot \mathbb{1}_{i>1} - \big[ 400 x_i (x_{i+1} - x_i^2) - 2(x_i - 1) \big] \cdot \mathbb{1}_{i<n}$, and $x^* = \{1, \dots, 1) \} \subset \{ x, \nabla g_n(x) = 0 \}$
\footnote{
When the coordinates range from $0$ to $n-1$, $g_n(x) = \sum_{i=0}^{n-2} \big[ 100 (x_{i+1} - x_i^2)^2 + (x_i - 1)^2 \big]$ and $\nabla_i g_n(x) = 200 (x_i - x_{i-1}^2) \cdot \mathbb{1}_{i>0} - \big[ 400 x_i (x_{i+1} - x_i^2) - 2(x_i - 1) \big] \cdot \mathbb{1}_{i<n-1}$.}. 
The number of stationary points of this function grows exponentially with dimensionality $n$, most of which are unstable saddle points \citep{10.1162/evco.2009.17.3.437}.

\begin{figure}[h!]
\centering
\includegraphics[width=13cm]{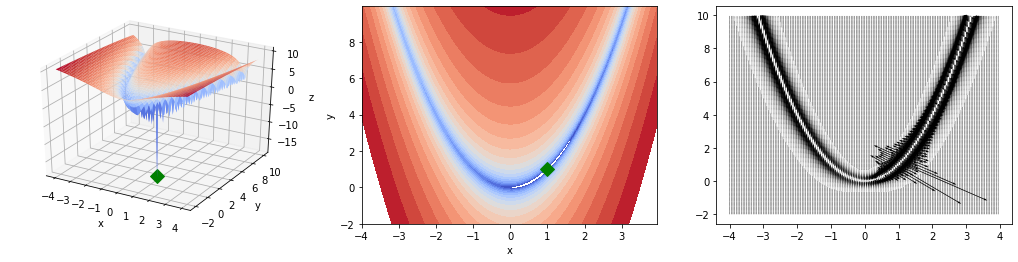}
\caption{Left) Rosenbrock function in log scale ($n=2$), Center) Contours, Right) Gradient field (note how this vector is pronounced in norm near the global minimum, which is important to understand why even near this global optimum many optimizers can felt to reach it)}
\label{fig:rosenbrock}
\end{figure}

We optimized the Rosenbrock function in a logarithmic scale (to create a ravine, figure \ref{fig:rosenbrock}). The function is unimodal, and the global minimum is very sharp and surrounded in the direction of the ravine by many local minima. At the beginning of optimization, we fall very quickly into the ravine because the surface is well-conditioned. Then, depending on the learning rate and the optimizer used (as well as the associated hyperparameters), we go down the ravine very slowly. Indeed, without momentum or velocity, we do not go directly down to the minimum since the gradient is almost zero along the ravine direction but very large in the perpendicular directions: we go from left to right (perpendicular to the ravine) while going down a little, but very slowly. Moreover,  we turn there almost indefinitely once we are near the minimum. With adaptive gradient, we go down to the minimum very quickly because this direction problem is corrected (due to momentum, left-right ravine perpendicular directions cancel out): if the learning rate is too small, we will also go down very slowly (small gradient in the flat ravine direction). Unlike SGD, here, we always reach the minimum (and stay there). Also, for some learning rates and initializations, there is a double descent \citep{DBLP:conf/iclr/NakkiranKBYBS20} in error (euclidean distance between the global minimum and the current position at a given time) when landing in the ravine.

\begin{figure}[h!]
\centering
\begin{subfigure}{.45\textwidth}
  \centering
  \includegraphics[width=1.0\linewidth]{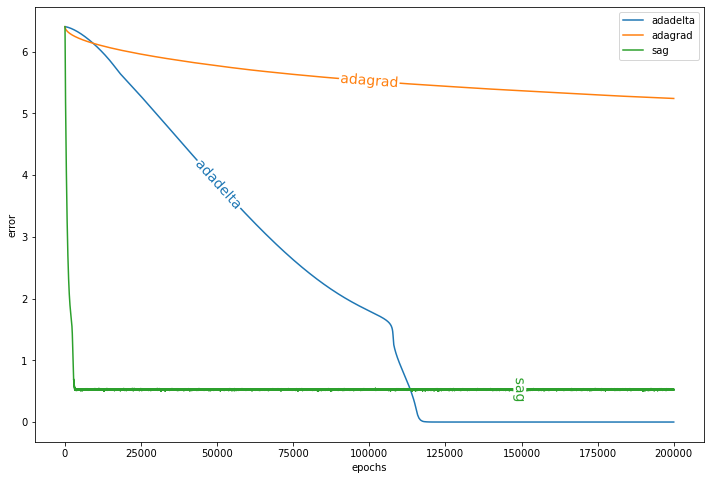}
  \caption{adadelta and adagrad vs sag}
  \label{fig:sag_vs_adagrad_vs_adadelta}
\end{subfigure}%
\begin{subfigure}{.45\textwidth}
  \centering
  \includegraphics[width=1.0\linewidth]{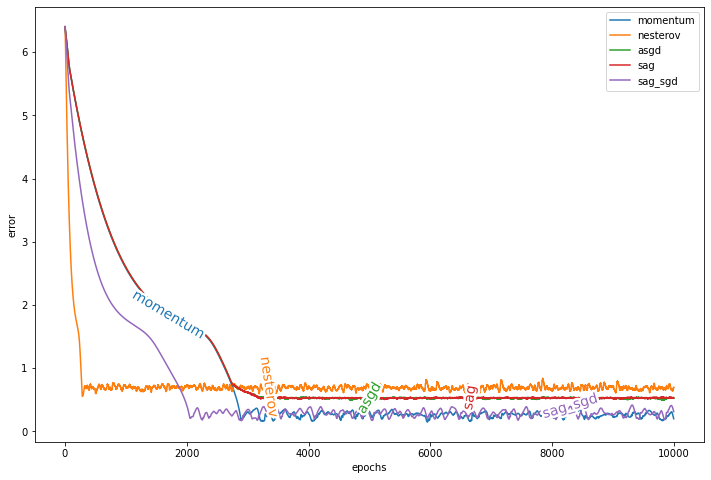}
  \caption{momentum, nesterov and asgd vs sag}
  \label{fig:sag_vs_sgd}
\end{subfigure} \\
\begin{subfigure}{.45\textwidth}
  \centering
  \includegraphics[width=1.0\linewidth]{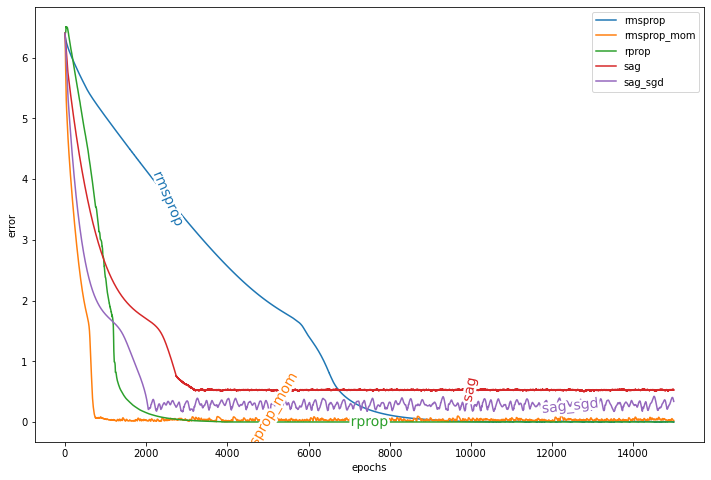}
  \caption{rmsprop, rmsprop\_mom and rprop vs sag}
  \label{fig:sag_vs_prop}
\end{subfigure}
\begin{subfigure}{.45\textwidth}
  \centering
  \includegraphics[width=1.0\linewidth]{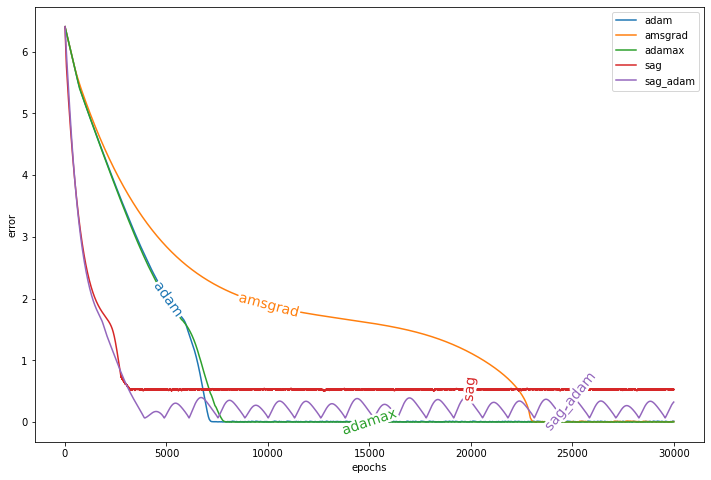}
  \caption{adam, amsgrad and adamax vs sag and sag\_adam}
  \label{fig:sag_vs_adam}
\end{subfigure}
\begin{subfigure}{.45\textwidth}
  \centering
  \includegraphics[width=1.0\linewidth]{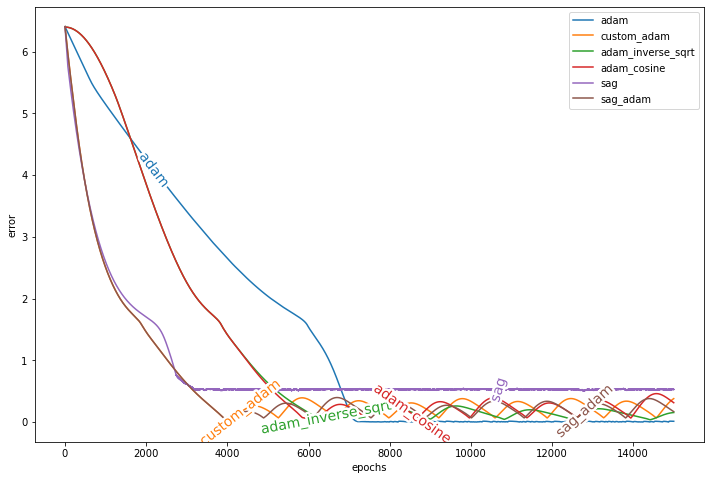}
  \caption{adam, custom\_adam, adam\_inverse\_sqrt, adam\_cosine}
  \label{fig:rosen_adam}
\end{subfigure}
\begin{subfigure}{.45\textwidth}
  \centering
  \includegraphics[width=1.0\linewidth]{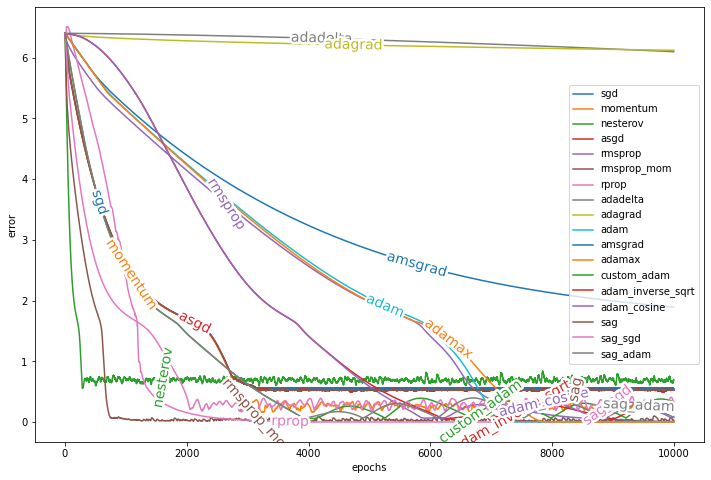}
  \caption{summary}
  \label{fig:sag_vs_all}
\end{subfigure}
\caption{Rosenbrock function}
\label{fig:rosenbrock}
\end{figure}

Adadelta and adagrad were very slow compared to sag. We can see in figure \ref{fig:sag_vs_adagrad_vs_adadelta} a comparative progression of these three algorithms. After 100 000 iterations adadelta and adagrad were still going down to the valley, while SAG did it in less than 1000 iterations, which is 100 times faster than both. Adadelta manages to reach the minimum, which sag never finally does.

Nosterov is faster on the well-conditioned part of the surface and arrives faster in the neighbourhood of the target than sag, momentum and asgd (figure \ref{fig:sag_vs_sgd}). On the other hand, it stabilizes at a higher loss than these methods. Sag and asgd have almost the same trajectory. Momentum follows the same trajectory as these two methods from the beginning but stabilizes at a smaller loss. The combination sag\_sgd (with momentum) speeds up the arrival in the neighbourhood of the minimum but stabilizes at the same level as momentum.

Rmsprop is slower than sag, but ends up with a smaller error than sag (figure \ref{fig:sag_vs_prop}). Adding momentum to rmsprop (rmsprop\_mom) improves its speed significantly. Rprop is also very fast and gives a smaller error than sag and sag\_sgd.

On the well-conditioned part of the surface, sag is faster than adam, adamax and amsgrad, but these methods reach the minimum (get zero final loss), which is not the case for sag (figure \ref{fig:sag_vs_adam}). The sag\_adam combination almost reaches the minimum, but is very chaotic and has periodic jumps that are similar to the slingshot mechanism \citep{https://doi.org/10.48550/arxiv.2206.04817}. amsgrad is much slower than adam and adamax.

custom\_adam, adam\_inverse\_sqrt, adam\_cosine also have the same periodic disruption phenomenon as sag\_adam (figure \ref{fig:rosen_adam}).

\begin{figure}[h!]
\centering
\includegraphics[width=15cm]{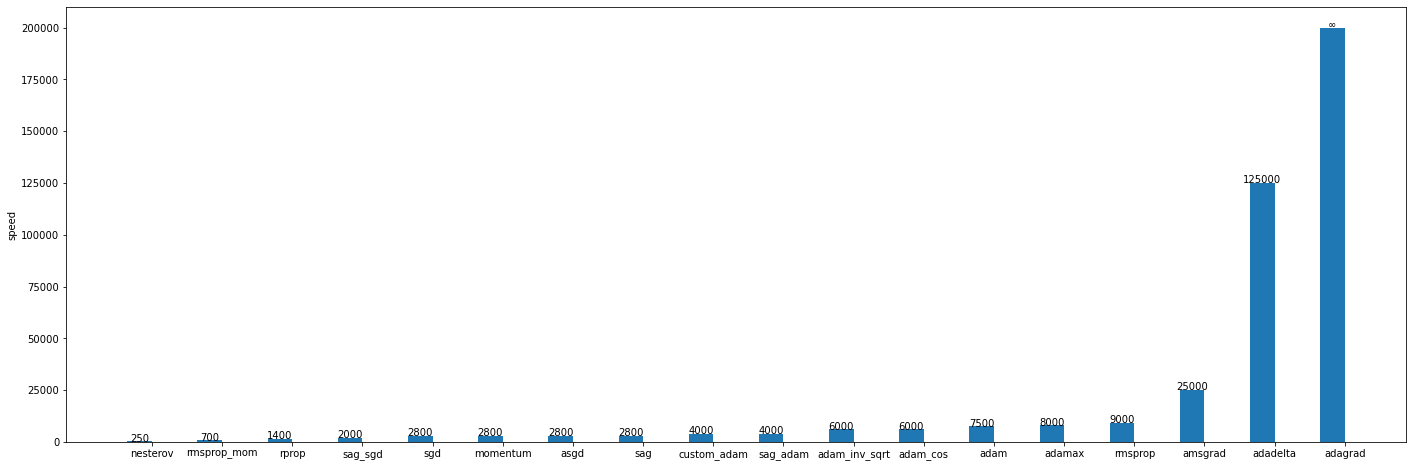}
\caption{Comparative visualization of convergence speeds on the rosenbrock function}
\label{fig:rosenbrock_speeds}
\end{figure}

\begin{figure}[h!]
\centering
\includegraphics[width=15cm]{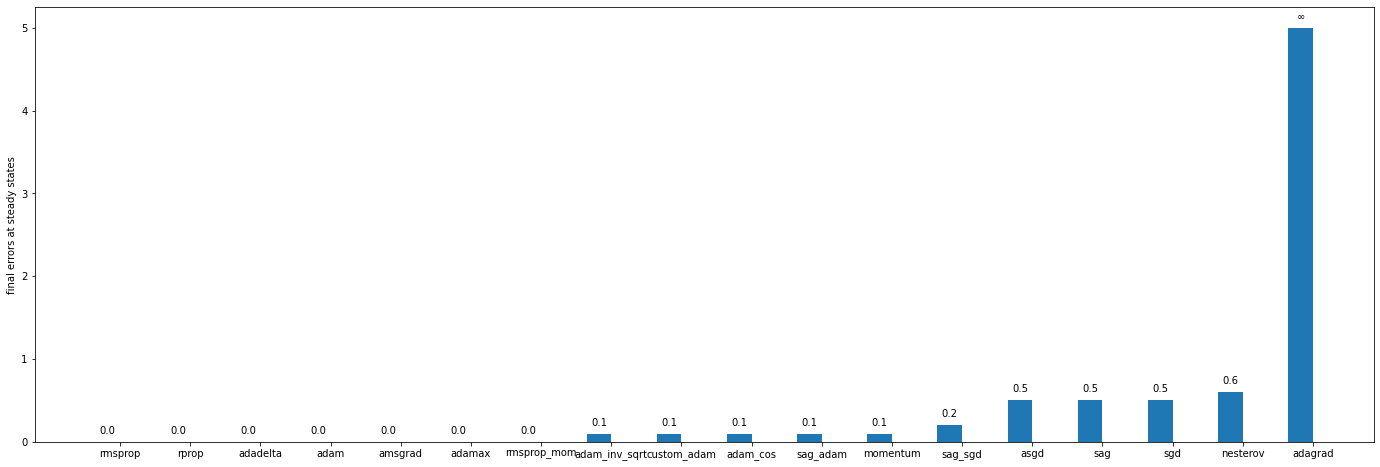}
\caption{Final errors at steady states on the rosenbrock function}
\label{fig:rosenbrock_errors}
\end{figure}

\begin{figure}[h!]
\centering
\includegraphics[width=15cm]{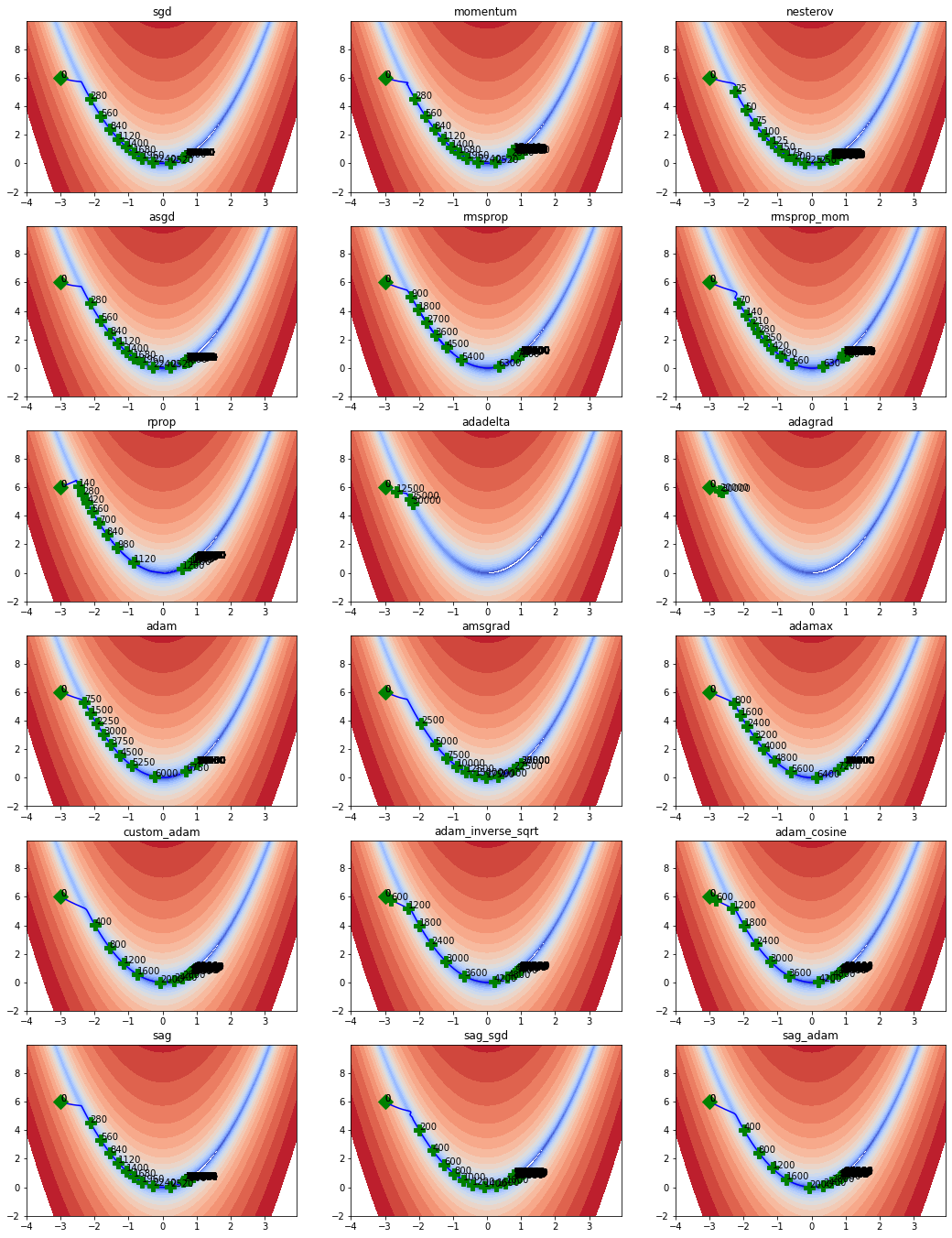}
\caption{Comparative visualization of the progression of each algorithm on the rosenbrock function}
\label{fig:rosen_progression}
\end{figure}

The methods that succeed in reaching the minimum are rmsprop, rprop, adadelta, adam, amsgrad,
adamax, rmsprop\_mom (figure \ref{fig:rosenbrock_errors}). The methods that come close to it without reaching it are adam\_inverse\_sqrt, custom\_adam, adam\_cosine, sag\_adam, momentum. The comparative convergence speeds are presented in figures \ref{fig:rosenbrock_speeds} and \ref{fig:rosen_progression}, which is an approximation of the number of iterations performed before reaching stabilization.

\subsubsection{Rastrigin function}


The rastrigin function is given by $g_n(x) = na + \sum_{i=1}^{n} \big[ x_i^2 - a \cos(2 \pi x_i)  \big] = n a + x^T x -  a 1_n^T \cos(2 \pi x)$ with $a \in \mathbb{R}$. Its gradient is $\nabla g_n(x) = 2x + 2 \pi a \sin (2\pi x)$, and $x^* = 
\{0, \dots, 0) \} \subset \{ x, \nabla g_n(x) = 0 \}$.

\begin{figure}[h!]
\centering
\includegraphics[width=13cm]{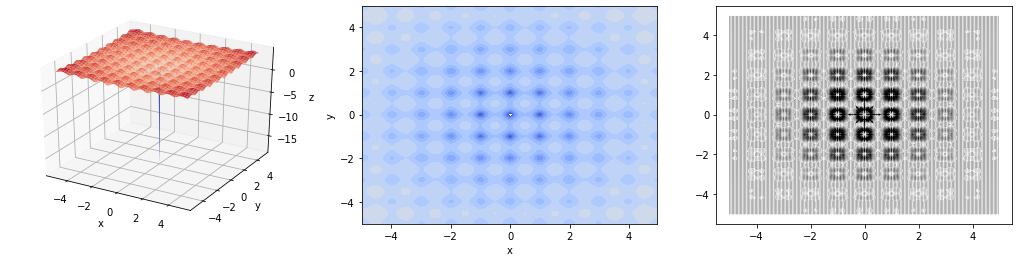}
\caption{Left) Rastrigin function in log scale ($A=10, n=2$), Center) Contours, Right) Gradient field}
\label{fig:rastrigin}
\end{figure}

We optimized the Rastrigin function in a logarithmic scale (to create many local minimums and make the global minimum sharp, figure \ref{fig:rastrigin}). The function is unimodal (in terms of global minimum), and the global minimum is very sharp and surrounded symmetrically by many local minima. At the beginning of optimization, we fall very quickly into the one local minimum. Then, depending on the learning rate and the optimizer used (and the associated hyperparameters), we can move successively from one minimum to another until we reach the global minimum.


\begin{figure}[h!]
\centering
\begin{subfigure}{.45\textwidth}
  \centering
  \includegraphics[width=1.0\linewidth]{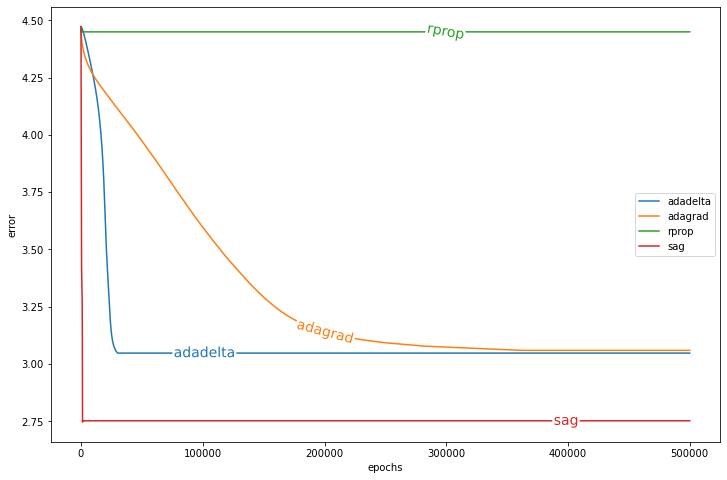}
  \caption{adadelta, adagrad and rprop vs sag}
  \label{fig:sag_vs_adagrad_vs_adadelta222}
\end{subfigure}%
\begin{subfigure}{.45\textwidth}
  \centering
  \includegraphics[width=1.0\linewidth]{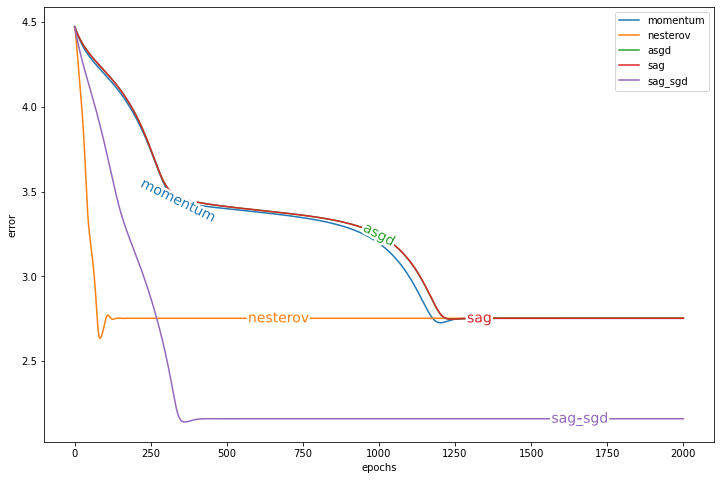}
  \caption{momentum, nesterov and asgd vs sag}
  \label{fig:sag_vs_sgd222}
\end{subfigure} \\
\begin{subfigure}{.45\textwidth}
  \centering
  \includegraphics[width=1.0\linewidth]{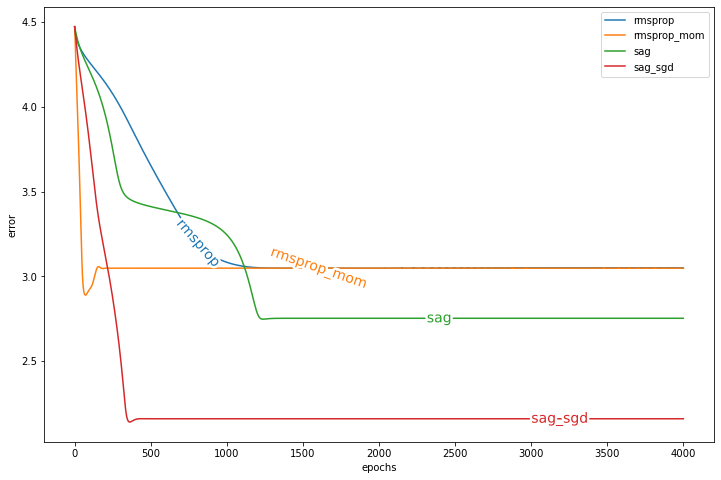}
  \caption{rmsprop, rmsprop\_mom  vs sag}
  \label{fig:sag_vs_prop222}
\end{subfigure}
\begin{subfigure}{.45\textwidth}
  \centering
  \includegraphics[width=1.0\linewidth]{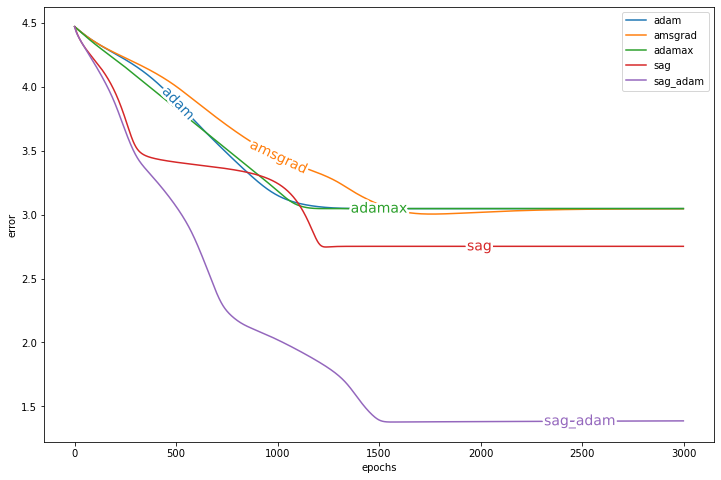}
  \caption{adam, amsgrad and adamax vs sag and sag\_adam}
  \label{fig:sag_vs_adam222}
\end{subfigure}
\begin{subfigure}{.45\textwidth}
  \centering
  \includegraphics[width=1.0\linewidth]{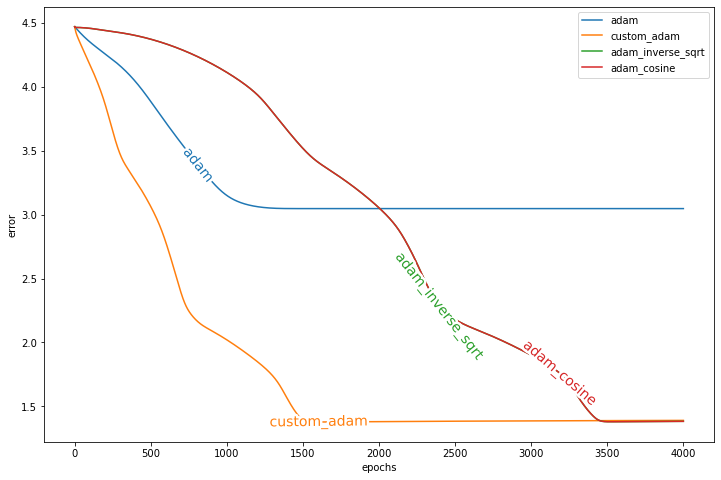}
  \caption{adam, custom\_adam, adam\_inverse\_sqrt, adam\_cosine}
  \label{fig:rosen_adam222}
\end{subfigure}
\begin{subfigure}{.45\textwidth}
  \centering
  \includegraphics[width=1.0\linewidth]{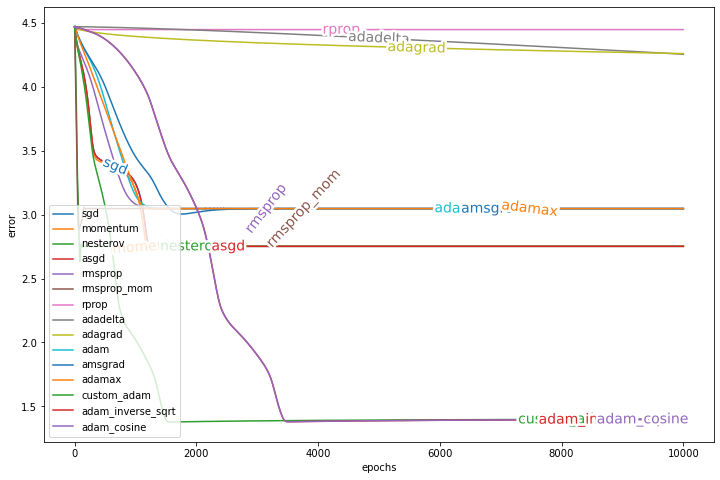}
  \caption{summary}
  \label{fig:sag_vs_all222}
\end{subfigure}
\caption{Rastrigin function}
\label{fig:rastrigin}
\end{figure}

Again, adadelta and adagrad are very slow compared to sag. We can see in figure \ref{fig:sag_vs_adagrad_vs_adadelta222} a comparative progression of these three algorithms. After 400 000 iterations adadelta and adagrad were still going down to the valley, while SAG did it in less than 1000 iterations, which is 400 times faster than both. adadelta manages to reach the minimum, which sag never finally does. Rprop is very bad here, it never leaves the first local minimum in which it falls. This is the method that obtains the largest error.

Nosterov is fast to reach the local minimum than sag, momentum and asgd (figure \ref{fig:sag_vs_sgd222}), and stabilizes at the same error as these methods. sag and asgd have almost the same trajectory. Momentum follows slightly the same trajectory as these two methods from the beginning and stabilizes at the same error. The combination sag\_sgd (with momentum) speeds up the arrival in the neighbourhood of the minimum and allows to obtain stabilization with a lower error. This means that it escapes more local minimums than the methods with which it is compared.

Rmsprop is slightly  slower than sag  and ends up with a bigger  error than sag (figure \ref{fig:sag_vs_prop222}). Adding momentum to rmsprop (rmsprop\_mom) improves its speed significantly, but we end up with the same error.

sag is faster than adam, adamax and amsgrad and gets a smaller error than them (figure \ref{fig:sag_vs_adam222}). The sag\_adam combination is much faster with less error. It is also one of the only methods to approach the global minimum (i.e. to escape so many obstacles). Amsgrad is much slower than adam and adamax, but ends up with the same error as them.

Custom\_adam is faster than adam\_inverse\_sqrt, adam\_cosine, but ends up with the same error as them (figure \ref{fig:rosen_adam222}).

\begin{figure}[h!]
\centering
\includegraphics[width=15cm]{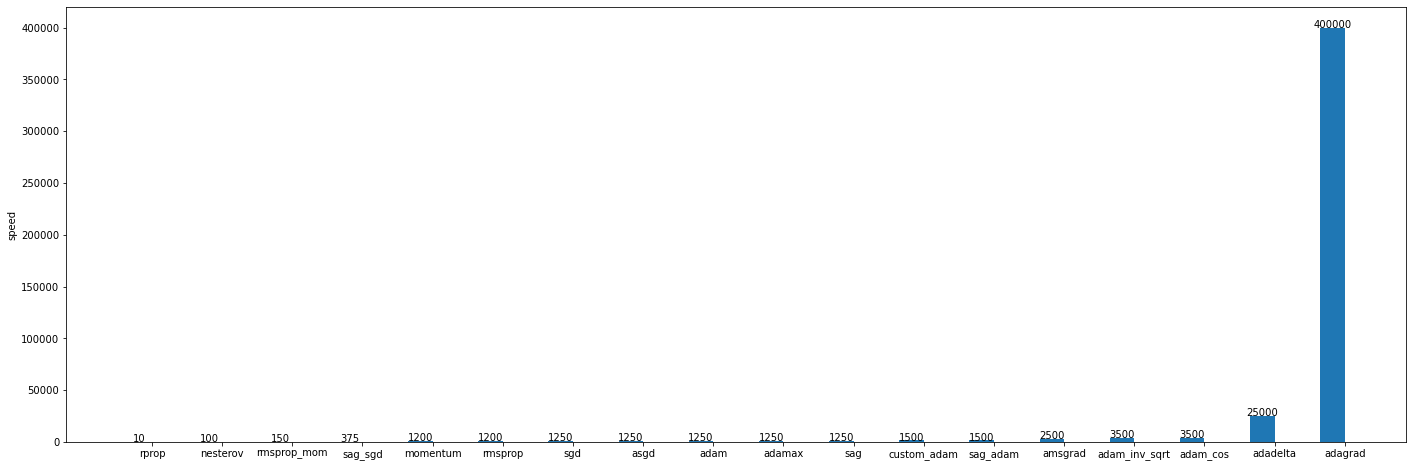}
\caption{Comparative visualization of convergence speeds on the rastrigin function}
\label{fig:rastrigin_speeds}
\end{figure}

\begin{figure}[h!]
\centering
\includegraphics[width=15cm]{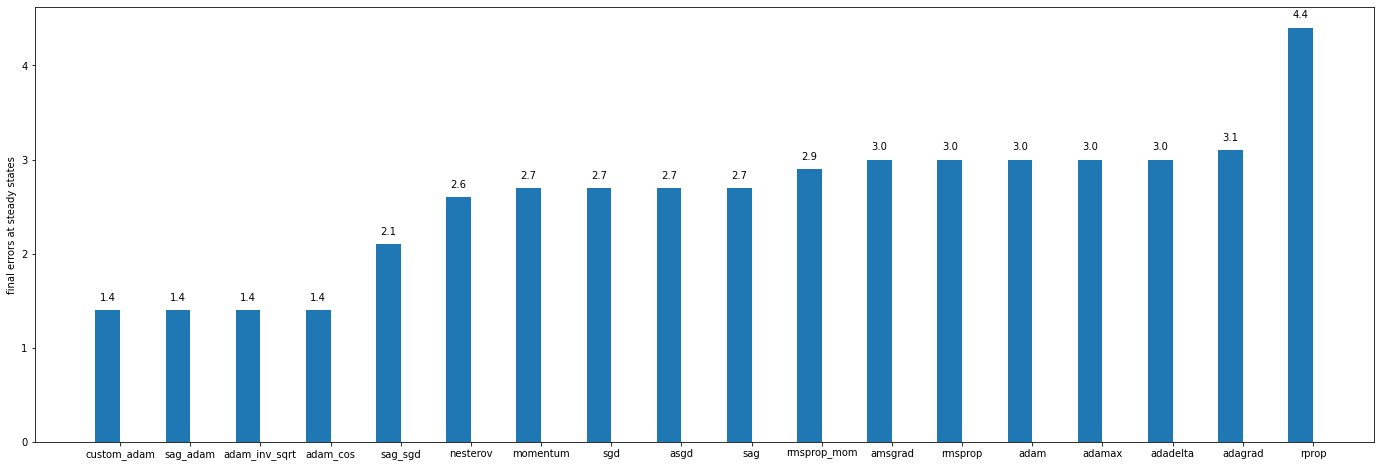}
\caption{Final errors at steady states on the rastrigin function}
\label{fig:rastrigin_errors}
\end{figure}

\begin{figure}[h!]
\centering
\includegraphics[width=15cm]{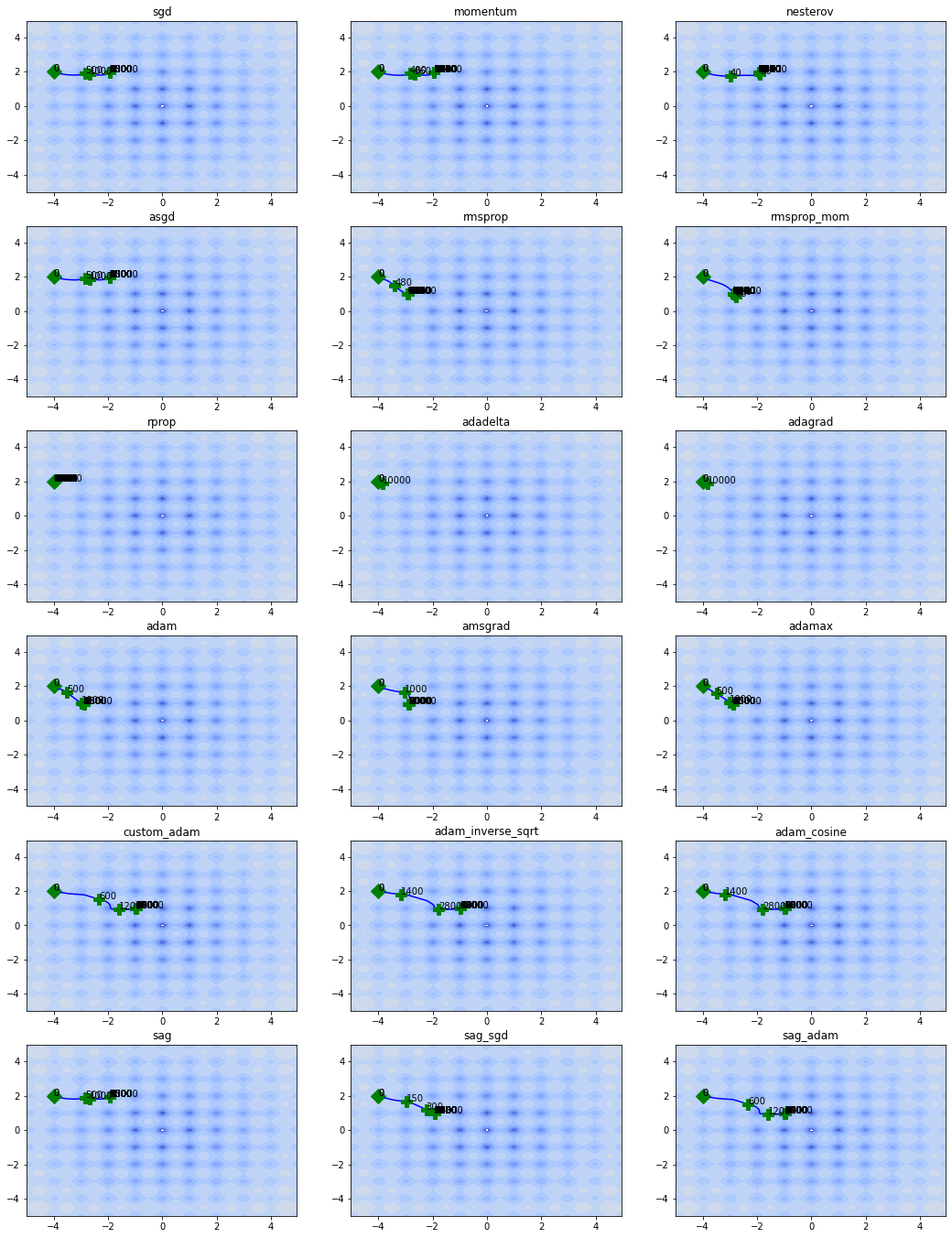}
\caption{Comparative visualization of the progression of each algorithm on the rastrigin function}
\label{fig:rastrigin_progression}
\end{figure}

No method has reached the global minimum (figure \ref{fig:rastrigin_errors}). The methods that come close to it without reaching it are adam\_inverse\_sqrt, custom\_adam, adam\_cosine and sag\_adam. The comparative convergence speeds are presented in figures \ref{fig:rastrigin_speeds} and \ref{fig:rastrigin_progression}, which approximate the number of iterations performed before reaching stabilization.

\subsection{Toys machine learning problems}

\subsubsection{Scikit-learn dataset}

We extracted the following datasets from scikit-learn \citep{scikit-learn}. The reader is 
invited to refer to the official scikit-learn website \footnote{\url{https://scikit-learn.org/stable/datasets/toy_dataset.html}} for more information about these data (sources, ...).
\begin{itemize}
    \item wine (classification): recognize the wine class given the features like the amount of alcohol, magnesium, phenol, colour intensity, etc.
    \item iris (classification): It contains sepal and petal lengths and widths for three classes of plants (Setosa, Versicolour, and Virginica)
    \item digits (classification): digit classification
    \item boston (regression): house prices in Boston based on the crime rate, nitric oxide concentration, number of rooms, distances to employment centers, tax rates, etc. The output feature is the median value of homes.
    \item diabete (regression): sklearn diabete dataset
    \item linnerud (regression): physical exercise Linnerud dataset
\end{itemize}

\begin{table}[h!]
\centering
\begin{tabular}{||P{2.0cm}|P{1.5cm}|P{1.5cm}|P{1.0cm}|P{2.5cm}|P{2.5cm}||}
\hline
Dataset & \# features & \# classes & size & train size (80\%) & val size (20\%) \\ 
\hline
wine & 13 & 3 & 178 & 142 & 36  \\ \hline
iris & 4 & 3 & 150 & 120 & 30  \\ \hline
digits & 64 & 10 & 1797 & 1437 & 360  \\ \hline
\end{tabular}
\newline \newline
\caption{Information about the sklearn datasets (classification)}
\label{tab:sklearn_ds}
\end{table}

\begin{table}[h!]
\centering
\begin{tabular}{||P{2.0cm}|P{1.5cm}|P{1.5cm}|P{1.0cm}|P{2.5cm}|P{2.5cm}||}
\hline
Dataset & \# features & \# output & size & train size (80\%) & val size (20\%) \\ 
\hline
boston & 13 & 1 & 506 & 404 & 102  \\ \hline
diabete & 10 & 1 & 442 & 353 & 89  \\ \hline
linnerud & 3 & 3 & 20 & 16 & 4  \\ \hline
\end{tabular}
\newline \newline
\caption{Information about the sklearn datasets (regression)}
\label{tab:sklearn_ds}
\end{table}

We trained a one-layer perceptron with a hidden layer of dimension 50, a leaky rectified linear unit (Leaky ReLU) activation (with a negative slope of 0.01) \citep{Maas2013RectifierNI} and a dropout of probability 0.1 \citep{JMLR:v15:srivastava14a}, this for 2000 epochs.

The results are presented in the following figures :
\begin{itemize}
\foreach \dataname in {wine, iris, digits, boston, linnerud, diabete}{
    \item \dataname : 
\foreach \figLabel in {adadelta_adagrad_sag,
momentum_nesterov_asgd_sag_sag_sgd,
adam_amsgrad_adamax_sag_sag_adam,
adam_custom_adam_adam_inverse_sqrt_adam_cosine_sag_sag_sgd_sag_adam,
all,
speeds}{
    \ref{fig:\dataname_\figLabel},
}
\ref{fig:\dataname_metrics}

}

\end{itemize}


\foreach \dataname in {wine, iris, digits, boston, linnerud, diabete}{

\begin{figure}[h!]
\centering
\includegraphics[width=17cm]{images/\dataname/adadelta_adagrad_sag.png}
\caption{adadelta, adagrad, sag (\dataname )}
\label{fig:\dataname_adadelta_adagrad_sag}
\end{figure}

\begin{figure}[h!]
\centering
\includegraphics[width=17cm]{images/\dataname/momentum_nesterov_asgd_sag_sag_sgd.png}
\caption{momentum, nesterov, asgd, sag, sag, sgd (\dataname)}
\label{fig:\dataname_momentum_nesterov_asgd_sag_sag_sgd}
\end{figure}

\begin{figure}[h!]
\centering
\includegraphics[width=17cm]{images/\dataname/adam_amsgrad_adamax_sag_sag_adam.png}
\caption{adam, amsgrad, adamax, sag, sag\_adam (\dataname )}
\label{fig:\dataname_adam_amsgrad_adamax_sag_sag_adam}
\end{figure}

\begin{figure}[h!]
\centering
\includegraphics[width=17cm]{images/\dataname/adam_custom_adam_adam_inverse_sqrt_adam_cosine_sag_sag_sgd_sag_adam.png}
\caption{adam, custom\_adam, adam\_inverse\_sqrt, adam\_cosine, sag, sag\_sgd, sag\_adam (\dataname )}
\label{fig:\dataname_adam_custom_adam_adam_inverse_sqrt_adam_cosine_sag_sag_sgd_sag_adam}
\end{figure}

\begin{figure}[h!]
\centering
\includegraphics[width=17cm]{images/\dataname/all.png}
\caption{Summary (\dataname )}
\label{fig:\dataname_all}
\end{figure}

\begin{figure}[h!]
\centering
\includegraphics[width=15cm]{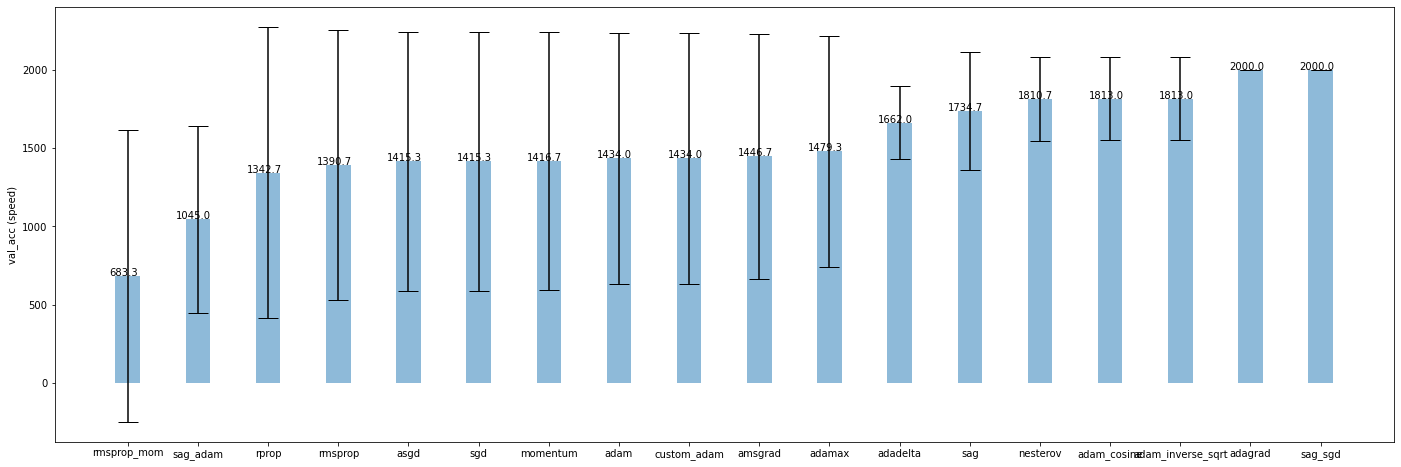}
\caption{Comparative visualization of convergence speeds (\dataname )}
\label{fig:\dataname_speeds}
\end{figure}

\begin{figure}[h!]
\centering
\includegraphics[width=15cm]{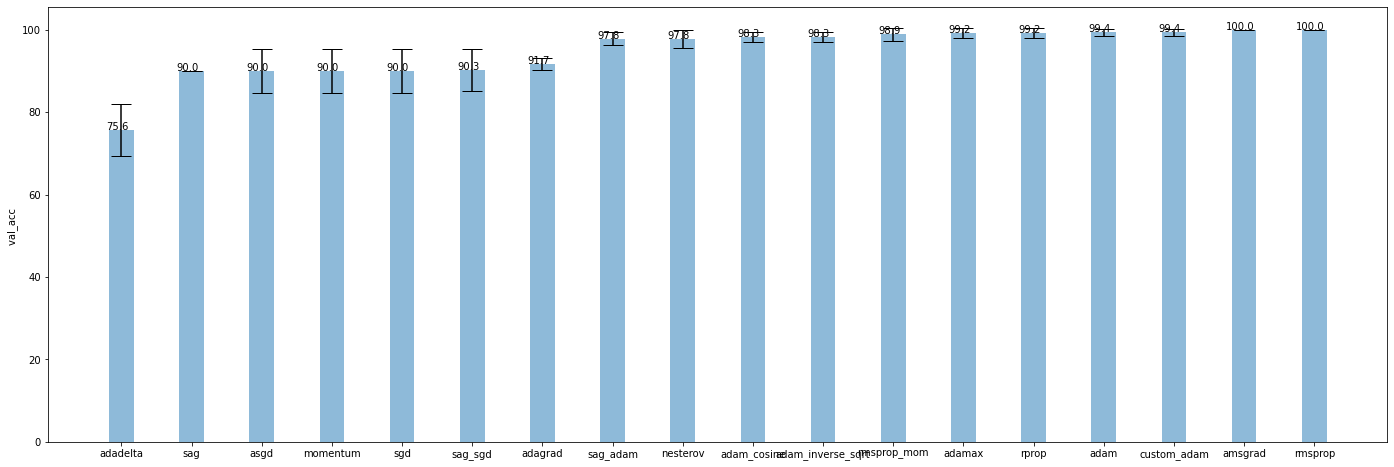}
\caption{Performances at steady states (\dataname )}
\label{fig:\dataname_metrics}
\end{figure}

}

\subsubsection{TorchVision dataset}

We extracted the datasets presented in table \ref{tab:torchvision_ds} from pytorch \citep{NEURIPS2019_9015}. The reader is  kindly invited to refer to the official pytorch website  \footnote{\url{https://pytorch.org/vision/stable/datasets.html}} for more information about these data (sources, ...).

\begin{table}[h!]
\centering
\begin{tabular}{||P{2.0cm}|P{3.8cm}|P{1.5cm}|P{1.0cm}|P{2.5cm}|P{2.5cm}||}
\hline
Dataset & (\# channels, height, width) & \# classes & size & train size (80\%) & val size (20\%) \\ 
\hline
mnist & (1, 28, 28) & 10 & 70000 & 60000 & 10000  \\ \hline
fashion mnist & (1, 28, 28) & 10 & 7000 & 60000 & 10000  \\ \hline
cifar10 & (3, 32, 32) & 10 & 60000 & 50000 & 10000  \\ \hline
cifar100 & (3, 32, 32) & 100 &  60000 & 50000 & 10000  \\ \hline
\end{tabular}
\newline \newline
\caption{TorchVision datasets}
\label{tab:torchvision_ds}
\end{table}

We trained a classifier having two main successive parts :
, and 

\begin{itemize}
    \item A first part consisting of two layers of convolutions neural networks :
    \begin{itemize}
        \item (0): Conv2d(\# channels, 10, kernel\_size=(5, 5), stride=(1, 1))
        \item (1): MaxPool2d(kernel\_size=2, stride=2, padding=0, dilation=1, ceil\_mode=False)
        \item (2): Conv2d(10, 10, kernel\_size=(5, 5), stride=(1, 1))
        \item (3): Dropout2d(p=0.1, inplace=False)
        \item (4): MaxPool2d(kernel\_size=2, stride=2, padding=0, dilation=1, ceil\_mode=False)
    \end{itemize}
    \item A second part consisting of a a two layers feed forward neural network : 
    \begin{itemize}
        \item (0): Linear(in\_features=160, out\_features=50, bias=True)
        \item (1): Dropout(p=0.1, inplace=False)
        \item (2): Linear(in\_features=50, out\_features=10, bias=True)
        \item (3): Dropout(p=0.1, inplace=False)
    \end{itemize}
\end{itemize}


\section{Summary and Discussion}
\label{sec:conclusion}

In this work, we compared the performance of SAG and several other  optimization algorithms for continuous objectives such as SGD with momentum, Nesterov Accelerated SGD, Averaged SGD, RMSProp (with and without momemtum), resilient backpropagation algorithm (Rprop), Adadelta, Adagrad, Adam, AMSGrad, Adamax, Adam with special learning rate decay procedure (inverse square root of the update number,  cyclical schedule that follows the cosine function).  SAG, although with a simple iteration, outperforms the majority of these algorithms. We have proposed two combinations of SAG. One with the momentum algorithm, which allows control of the importance of each gradient term in the mean used by SAG depending on the iteration during which it is used, and another with Adam where the importance of the square of the norm of the gradient is also controlled. These two variants allowed us to improve the speed empirically while obtaining better performances.

\paragraph{Limitations} The memory cost used by SAG is very high compared to other algorithms, which makes it impractical for large scale use.

\paragraph{Perspectives} What we presented as an improvement is only an empirical illustration of the performance of SAG. It would be interesting to evaluate theoretically the expected convergence rate of all these algorithms. We leave this for future work.

\section*{Acknowledgement}

The authors thank Fabian Bastin who made this work possible, and for discussion at the early stage of this project during the stochastic programming (IFT6512) course at UdeM (Université de Montréal). We also thank Compute Canada for computational resources.

\bibliography{neurips_2021}
\bibliographystyle{neurips_2021}

\appendix



\end{document}